\documentclass{article}
\usepackage{ijcai21}

\usepackage{times}
\usepackage{soul}
\usepackage{url}
\usepackage[hidelinks]{hyperref}
\usepackage[utf8]{inputenc}
\usepackage[small]{caption}
\usepackage{subcaption}
\usepackage{graphicx}
\usepackage{amsmath}
\usepackage{amssymb}
\usepackage{amsthm}
\usepackage{booktabs}
\usepackage{algorithm}
\usepackage{algorithmic}
\usepackage{comment}
\usepackage{graphicx}
\usepackage{multirow}
\usepackage{tikz}
\usetikzlibrary{arrows, positioning}
\usepackage{flowchart}
\usepackage{hyphenat}
\usepackage[nomain, acronym, nonumberlist]{glossaries}
\usepackage{csquotes}
\usepackage{color}
\usepackage{booktabs}
\usepackage{subcaption}
\usepackage{titlesec}

\newcommand{\eg}{e.g., } 
\newcommand{\ie}{i.e., }
\newcommand{\important}[1]{\textit{#1}}

\newcommand{\model}{SimDem }
\newcommand{\pwd}{PwD}

\tikzstyle{io} = [trapezium, trapezium left angle=70, trapezium right angle=110, minimum width=1cm, minimum height=1cm, text centered, draw=black]
\glsaddkey*
	{domain} 
	{\glsentrytext{\glslabel}domain} 
	{\glsentrydomain}
	{\Glsentrydomain}
	{\glsdomain}
	{\Glsdomain}
	{\GLSdomain}

\newglossarystyle{formaldef}{
	
	\renewcommand*{\glsgroupheading}[1]{}
	
}

\newenvironment{myitemize}
{ \begin{itemize}
    \setlength{\itemsep}{0.75pt}
    \setlength{\parskip}{0pt}
    \setlength{\parsep}{0pt}     }
{ \end{itemize}                  }  

\newglossary*{patient}{Header to be removed}
\newglossary*{nurse}{formal definition of a nurse type agent}
\newglossary*{watch}{formal definition of an SAMi-Watch type agent}
\newglossary*{need}{formal definition of a need}
\newglossary*{map}{formal definition of a cognitive map}
\newglossary*{domain}{domains and notational conventions}
\newglossary*{environment}{formal definition of the environment}

\newacronym{app}{APP}{amyloid precursor protein}
\newacronym{apoe4}{APOE4}{apolipoprotein E4}
\newacronym{ad}{AD}{Alzheimer's disease}
\newacronym{mci}{MCI}{mild cognitive impairment}
\newacronym{adl}{ADL}{activities of daily living}
\newacronym{pwd}{PwD}{people with dementia}
\newacronym{actr}{ACT-R}{Adaptive Control of Thought-Rationale}
\newacronym{bdi}{BDI}{belief-desire-intention}
\newacronym{ftd}{FTD}{frontotemporal dementia}
\newacronym{lda}{LDA}{linear discriminant analysis}
\newacronym{gps}{GPS}{Global Positioning System}
\newacronym{sami}{SAMi}{Sensor-based individualised activity management system for people with dementia in nursing facilities} 
\newacronym{pmf}{PMF}{probability mass function}
\newacronym{ble}{BLE}{Bluetooth Low Energy}
\newacronym{png}{PNG}{Portable Network Graphics}
\newacronym{json}{JSON}{JavaScript Object Notation}
\newacronym{dda}{DDA}{Digital Differential Analyzer}

\newglossaryentry{need:requested}{
	name={requested features},
	symbol=\ensuremath{R},
	domain=\ensuremath{\subseteq \glssymbol{env:features}},
	description={is the set of grid cell features requested by this need in order to satisfy it (\eg $\text{\enquote{toilet}}$)}}

\newglossaryentry{need:state}{
	name={is active},
	symbol=\ensuremath{q_N},
	domain=\ensuremath{\in \{q_0,q_1\}},
	description={is the state of the need that specifies, whether the need is currently active ($q_1$) or not ($q_0$).}}
	
\newglossaryentry{need:fstep}{
	name={step function},
	symbol=\ensuremath{f_N},
	domain=\ensuremath{: \glssymbol{dom:pat} \mapsto \{q_0, q_1\}},
	description={is the step function that determines if the need will be active in the next simulation step.}}

\newglossaryentry{map:workingmemory}{
	name={working memory},
	symbol=\ensuremath{M},
	domain=\ensuremath{\subseteq \glssymbol{env:cells}},
	description={is a set of grid cells and describes the agents \important{working memory}. As the agent moves around, perceived grid cells will be automatically added to the working memory.}}
	
\newglossaryentry{map:capacity}{
	name={capacity},
	symbol=\ensuremath{c},
	domain=\ensuremath{\in \mathbb{N}},
	description={is the \important{capacity} of the agents cognitive map, \ie the maximum number of grid cells that the agent can remember at the same time. Once the capacity is exceeded, grid cells will be removed from the working memory in the order they were added.}}

\newglossaryentry{map:pforgetcell}{
	name={cell forget probability},
	symbol=\ensuremath{p_c},
	domain=\ensuremath{\in [0,1]},
	description={denotes the probability per simulation step, that a currently stored grid cell will be forgotten and removed from the working memory.}}

\newglossaryentry{pat:position}{
	name=position,
	symbol=\ensuremath{x},
	domain=\ensuremath{\in \glssymbol{env:cells}},
	description={is the \important{position} of the PwD type agent on the cell grid.}}

\newglossaryentry{pat:home}{
	name={home features},
	symbol=\ensuremath{h},
	domain=\ensuremath{\subseteq\glssymbol{env:features}},
	description={is the set of features, that defines the PwD's \important{home}, \ie his personal room.}}

\newglossaryentry{pat:schedule}{
	name=schedule,
	symbol=\ensuremath{S},
	domain=,
	description={is \important{daily schedule} of the PwD type agent and is a set of appointments}}
	
\newglossaryentry{pat:pforgetappointment}{
	name={schedule forgetting probability},
	symbol=\ensuremath{p_{\glssymbol{pat:schedule}}},
	domain=\ensuremath{\in [0,1]},
	description={is the probability that a PwD might forget an appointment.}}

\newglossaryentry{pat:needs}{
	name={needs},
	symbol=\ensuremath{N},
	domain=\ensuremath{=(n_0,n_1,\ldots,n_k)},
	description={is a tuple of \important{needs}, that the PwD type agent may experience and include physiological (\eg sleep) and psychological (\eg socialising) desires along with the need to meet appointments. Appointments are represented in form of a daily schedule ($S$) that contains the time and location of each appointment. Whereas \glssymbol{pat:pforgetappointment} \glsdomain{pat:pforgetappointment} is the probability that a PwD might forget an appointment.}}
	
\newglossaryentry{pat:cmap}{
	name={cognitive map},
	symbol=\ensuremath{c},
	domain=\ensuremath{},
	description={represents agent's \important{cognitive map}}}

\newglossaryentry{pat:state}{
	name={cognitive state},
	symbol=\ensuremath{q},
	domain=\ensuremath{\in \{q_O, q_D, q_G\}},
	description={is the agent's \important{cognitive state}, \ie if PwD is oriented ($q_O$), disoriented ($q_D$) or currently guided by a nurse type agent ($q_G$).}}

\newglossaryentry{pat:sight}{
	name={sight},
	symbol=\ensuremath{s},
	domain=\ensuremath{\in \mathbb{R}^+},
	description={is the PwD's  \important{sight}, \ie the radius of the circle sector that represents his attention zone.}}

\newglossaryentry{pat:fov}{
	name={field of view},
	symbol=\ensuremath{\alpha_{FOV}},
	domain=\ensuremath{\in \mathbb{R}^+},
	description={is the PwD's  \important{field of view}, \ie the angle of the circle segment that represents his attention zone.}}

\newglossaryentry{pat:plandmarks}{
	name={landmark function},
	symbol=\ensuremath{f_{\lambda}},
	domain=\ensuremath{: \glssymbol{env:features} \mapsto [0,1]},
	description={is the \important{landmark function}, that assigns each landmark a probability that the agent will regain orientation after perceiving a grid cell of the given feature.}}

\newglossaryentry{pat:pinterventions}{
	name={intervention probability},
	symbol=\ensuremath{f_{i}},
	domain=\ensuremath{: \{ i_R, i_N \} \mapsto [0,1]},
	description={is the \important{intervention function} and specifies the probability that a reminder or navigation intervention triggered by a smart watch is successful.}}

\newglossaryentry{nur:position}{
	name=position,
	symbol=\ensuremath{x},
	domain=\ensuremath{\in \glssymbol{env:cells}},
	description={is the \important{position} of the nurse type agent on the cell grid.}}

\newglossaryentry{nur:state}{
	name={state},
	symbol=\ensuremath{q},
	domain=\ensuremath{\in \{q_I, q_P, q_G\}},
	description={is the nurse's \important{state}, \ie if the nurse is currently idling ($q_I$), pursuing a disoriented PwD ($q_P$) or guiding a PwD to his destination ($q_G$).}}

\newglossaryentry{nur:home}{
	name={home},
	symbol=\ensuremath{h},
	domain=\ensuremath{\subseteq\glssymbol{env:features}},
	description={is the set of features, that defines the nurse's \important{home}, \ie the features of grid cells that the nurse will attend while idling.}}

\newglossaryentry{nur:sight}{
	name={sight},
	symbol=\ensuremath{s},
	domain=\ensuremath{\in \mathbb{R}^+},
	description={is the nurse's \important{sight}, which is a complete circle contrary to PwD type agent.}}

\newglossaryentry{nur:patient}{
	name=patient,
	symbol=\ensuremath{a_p},
	domain=\ensuremath{\in \{ \varepsilon \} \cup \glssymbol{dom:pat}},
	description={is the slot for a PwD type agent that currently needs help from the nurse.}}

\newglossaryentry{watch:patient}{
	name=patient,
	symbol=\ensuremath{a_p},
	domain=\ensuremath{\in \glssymbol{dom:pat}},
	description={is the PwD agent monitored by the smart watch.}}

\newglossaryentry{watch:state}{
	name={state},
	symbol=\ensuremath{q},
	domain=\ensuremath{\in \{q_M, q_W\}},
	description={is the \important{state} of the smart watch, \ie if it is currently monitoring the PwD ($q_M$) or waiting for a nurse to help the PwD ($q_W$).}}
	
\newglossaryentry{watch:cooldown}{
	name={cooldown time},
	symbol=\ensuremath{\tau_{cd}},
	domain=\ensuremath{\in \mathbb{R}^+},
	description={is the \important{cooldown time}, \ie the amount of time in seconds, that has to pass after an unsuccessful intervention, before another intervention will be attempted.}}

\newglossaryentry{watch:sensormodel}{
	name={sensormodel},
	symbol=\ensuremath{p_{d}},
	domain=\ensuremath{\in [0,1]},
	description={represents the \important{sensor model} of the  smart watch and denotes the probability for each simulation step that disorientation of the monitored PwD is detected by the smart-Watch.}}

\newglossaryentry{watch:nhelp}{
	name={maximum number of failed intervention},
	symbol=\ensuremath{n_{help}},
	domain=\ensuremath{\in \mathbb{N}^{*}},
	description={denotes the maximum number of consecutive, failed interventions (\ensuremath{n} \ensuremath{\in \mathbb{N}}), before a help intervention is taken.}}
	
\newglossaryentry{watch:nfail}{
	name={number of failed intervention},
	symbol=\ensuremath{n},
	domain=\ensuremath{\in \mathbb{N}},
	description={is the number of consecutive, failed interventions.}}

\newglossaryentry{watch:tnext}{
	name={time of next available intervention},
	symbol=\ensuremath{\tau_{next}},
	domain=\ensuremath{\in \glssymbol{dom:time}},
	description={denotes the time point where the next intervention can be made.}}

\newglossaryentry{dom:time}{
	name=time,
	symbol=\ensuremath{D_T},
	domain=\ensuremath{= \{t_0, t_1, \ldots, t_k\}},
	description={is the \important{time domain}, that contains all time points of the simulation.}}

\newglossaryentry{dom:pat}{
	name={PwD type agents},
	symbol=\ensuremath{D_p},
	domain=\ensuremath{= \{p_0, p_1, \ldots, p_k\}},
	description={is the set of PwD type agents present in the model.}}
	
\newglossaryentry{dom:nur}{
	name={nurse type agents},
	symbol=\ensuremath{D_n},
	domain=\ensuremath{= \{n_0, n_1, \ldots, n_k\}},
	description={is the set of nurse type agents present in the model.}}

\newglossaryentry{dom:wat}{
	name={SAMi-Watch type agents},
	symbol=\ensuremath{D_w},
	domain=\ensuremath{= \{w_0, w_1, \ldots, w_k\}},
	description={is the set of SAMi-Watch type agents present in the model.}}

\newglossaryentry{dom:null}{
	name={null element},
	symbol=\ensuremath{\varepsilon},
	domain=\ensuremath{},
	description={denotes a special element that represents a null type object.}}
\newglossaryentry{env:cells}{
	name=cells,
	symbol=\ensuremath{V},
	domain=\ensuremath{=[0,\ldots,n-1]\times[0,\ldots,m-1]},
	description={is the set of \important{grid cells}, where $n$ is the width and $m$ is the height of the grid.}}

\newglossaryentry{env:nhood}{
	name=neighbourhood,
	symbol=\ensuremath{N},
	domain=\ensuremath{=\{(-1,-1), (-1,0), (-1,1), (0,-1), (0,1), (1,-1), (1,0), (1,1)\}},
	description={is the \important{neighbourhood} of the grid cells. Here, it is a Moore neighbourhood.}}
	
\newglossaryentry{env:features}{
	name=properties,
	symbol=\ensuremath{\Sigma},
	domain=\ensuremath{= \{\sigma_0, \sigma_1, \ldots, \sigma_l\}},
	description={is the set of existing \important{grid cell features}.}}
	
\newglossaryentry{env:featuremap}{
	name={feature map},
	symbol=\ensuremath{f_{\sigma}},
	domain=\ensuremath{: \glssymbol{env:cells} \mapsto 2^{\glssymbol{env:features}}},
	description={is the \important{feature map}, that assigns a set of features to each grid cell.}}

\makenoidxglossaries

\title{SimDem -- A Multi-agent Simulation Environment to Model Persons with Dementia and their Assistance}

\author{
Muhammad Salman Shaukat$^1$
\and
Bjarne Christian Hiller$^1$\and
Sebastian Bader$^1$\And
Thomas Kirste$^1$
\affiliations
$^1$Department of Computer Science, University of Rostock, Rostock, Germany
}

\begin{document}

\maketitle
\begin{abstract}
  Developing artificial intelligence based assistive systems to aid Persons with Dementia (\pwd) requires large amounts of training data. However, data collection poses ethical, legal, economic, and logistic issues. Synthetic data generation tools, in this regard, provide a potential solution. However, we believe that already available such tools do not adequately reflect cognitive deficiencies in behavior simulation. To counter these issues we propose a simulation model (\model) that primarily focuses on cognitive impairments suffered by PwD and can be easily configured and adapted by the users to model and evaluate assistive solutions. 
  \end{abstract}

\section{Introduction}
Dementia refers to various symptoms that result in impairment of memory, critical thinking, ability to conduct simple daily tasks. Dementia has become a major concern worldwide, and puts a significant burden on the healthcare sector. Emerging assistive technologies have the potential to relieve caregiver's burden, improve quality-of-life of \pwd, and reduce overall cost associated with the caregiving process. However, to assess the usefulness of these assistive technologies, clinical experiments involving PwD are required which pose ethical, legal, economical, and logistic concerns related to the data collection process. In this regard, a simulation model that could generate synthetic data on \pwd's navigation behavior and their response towards assistance could bring a potential solution for model building and validation.

A critical aspect for modeling \pwd's behavior is their spatial navigation, which is an aggregated task of \textit{wayfinding} and \textit{locomotion}. Wayfinding refers to the  \textit{allocentric} cognitive element of the navigation task that involves tactical and strategic planning to guide the movement (e.g., goal selection, route determination). Whereas locomotion refers to the \textit{egocentric} physical process of the movement to reach a location (e.g., walking, taking a turn) \cite{DP14,gunzelmann2006mechanisms}.

Cognitive competence associated with egocentric and allocentric strategies can be represented as a \textit{cognitive map}, a term first introduced by \citeauthor{tolman_cognitive_1948} (1948). The cognitive map is a mental representation of an environment. It is hypothesized that spatial information of the environment (\eg landmarks, goals, boundaries etc.) is encoded within \textit{spatial cells} found in our \textit{hippocampus} that guides our spatial orientation \cite{hartley2014space}. Hence, we represent cognitive map as a combination of spatial grid cells that encode spatial semantics of the environment such as goal area, boundaries, landmarks etc. Moreover, PwD show a decline in spatial competence as the disease progresses, this decline is believed to be  linked with the ability to \enquote{memorize} such a cognitive map \cite{coughlan2018spatial}. We utilize this concept as the \textit{cognitive capacity} to quantify spatial competence. Based on mentioned cognitive principles, in this article, we present \textbf{\model}: A Multi-agent \textbf{Sim}ulation Environment to Model Persons with \textbf{Dem}entia and their Assistance for indoor environments.

\section{Related Work}
\citeauthor{huang_indoorstg_2013} (2016) proposed the IndoorSTG tool to generate trajectories of customers and assistants in a shopping mall based on random movement for customers and location-likelihood based probabilistic movement for assistants. \citeauthor{li_vita_2016} (2016) proposed the Vita tool that takes a more microscopic approach and allows movement customization. \citeauthor{ryoo_generation_2016} (2016) pointed out the lack of movement customization and semantic indoor context in IndoorSTG and Vita and introduced walking speeds based on location, movement constraints (\eg boundaries) and mobility settings for each individual.

However, these tools do not consider cognitive impairment and only consider healthy individuals. Tailoring these simulation systems to model PwD is either infeasible (\eg producing random walk) or not possible at all (\eg availability). Therefore, we investigated computational models that account for \textit{wayfinding uncertainty}, to incorporate them in our simulation architecture, that refers to the way-finder's uncertainty over choosing correct path, and goal etc.   \citeauthor{griffith_simulation_2017} (2017) proposed a model to simulate human wayfinding that focused on quantifying effects of dementia by introducing agent-based parameters such as attention zone (\ie radius, arc length) and spatial orientation abilities through landmark recognition. \citeauthor{manning_magellan_2014} (2014) proposed a cognitive map based spatial learning mechanism, MAGELLAN, in their taxicab driving  simulation. To correlate cognitive ability with wayfinding uncertainty, they utilized cognitive parameters \textit{vision} and \textit{memory}. \citeauthor{andresen_wayfinding_2016} (2016) presented a wayfinding model using the concept of cognitive map along with semantic spatial knowledge (\eg common rooms and circulation rooms). \citeauthor{kielar_unified_2018} (2017) modeled spatial and social cognition associated with wayfinding by proposing routing methods to model navigation based on perfect knowledge and uncertain knowledge of the environment. \citeauthor{zhao_multi-strategy_2011} (2011) proposed a egocentric and allocentric based navigation model based on cognitive architecture, Adaptive Control of Thought-Rationale (ACT-R). Majority of reviewed literature only accounted for healthy individuals, however, spatial knowledge representation as cognitive maps provides a potential solution to model navigation behavior for PwD. In addition to model based approaches, \textit{interactive approaches} can provide a more comprehensive and controlled activity datasets, however they require user-interaction and thus limits extensive data generation \cite{Synnott2015}. Therefore, we chose a model based approach to design the simulation system presented in this work.

\section{The \model model}
\label{par:case1}

\subsection{Requirements}
\label{sec:requirements}
To identify the simulation model requirements, let us consider a case-study: Mr. Alois is a permanent resident at a nursing home who is provided with a personal smart watch that assists him in managing his daily tasks. He starts his day with breakfast at 8:00 and has a scheduled medical examination that he forgot. The smart watch gives a reminder notification to him which happens to be successful and he reaches the examination spot 5 minutes later. Later that day, Mr. Alois goes for lunch at 12:00 on his own without any assistance. At 15:00, he has physiotherapy, but despite the navigational aid of the smart watch, he gets lost constantly. Upon observing his behavior, the smart watch eventually calls the nursing staff with his location. The nurse then reaches Mr. Alois and accompanies him to the physiotherapy appointment. Other than scheduled appointments, Mr. Alois satisfies his physiological (e.g., sleep, toilet) and social (e.g., meeting with other residents) needs. Spatial landmarks such as signs, picture on wall, often help Mr. Alois during navigation. The case-study illustrates that a successful smart watch based intervention can improve PwD's independence and decrease the amount of human resources being spent on the PwD. Therefore, an important requirement of the simulation system is to incorporate the smart-watch based assistance which can be later used to evaluate different intervention strategies that might help to improve the efficiency of such assistance. Further requirements are as follows: \\
\textbf{Environment:}
The Simulation model must incorporate a spatial representation of the environment to represent spatial information of the environment such as places (\eg clinic, dining area, toilets), spatial landmarks, and boundaries (\eg walls and obstacles).\\
\textbf{Agents:}
The Model should be able to simulate different types of agents with different abilities and roles. We focus on three main types of agents: 1) PwD agent, 2) Nurse agent, 3) Smart-watch agent. Moreover, different types of agent should be able to interact where the type of interaction depends on the context. For instance, the smart watch agent should be able to communicate with the PwD agent (e.g., issuing a reminder for an appointment) and with the nursing agent (e.g., notifying the nurse if the PwD agent needs a physical intervention). Furthermore, The nurse agent should have the ability to takeover the PwD agent (i.e., the PwD agent should follow the nurse agent) when required. Ideally, the simulation model should incorporate the concept of non-compliance in certain situations where the PwD agent might refuse to follow a fixed schedule or might not cooperate with the nursing staff. However, our current model assumes that PwD agents are always cooperative with others and do not oppose their schedule. Moreover, human type agents should be able to perceive the environment such as a PwD agent visualizing landmarks or nurse agent perceiving a disoriented PwD agent.\\
\textbf{Schedule and Needs:}
To plan a PwD's day or appointments a schedule is required that can be followed by the PwD agent. Moreover, PwD's physiological (e.g., sleep, toilet) and psychological (e.g., socializing) needs should also be incorporated.

\subsection{\model Concept}
\label{sec:concept}
\paragraph{Environment:}
To model an environment, continuous coordinate systems enable precise modeling of spatial relationships. However, due to their computation complexity, pedestrian behavior simulations adapt space discretization techniques to reduce computational expensiveness. As identified in \citeauthor{nasir_prediction_2014}(2014), we represent environment as a two-dimensional orthogonal grid of size $n\times m$. Grid cells are connected via \textit{Moore neighborhood} with a side length of $50 cm$ for each cell. To avoid collision among agents, each cell can be occupied by a single agent a time. Each grid cell encodes spatial features such as locations (\eg living room, clinic, toilet etc), boundaries, and spatial landmarks.  From the cognitive perspective, grid map with spatial information forms the basis for PwD type agent's cognitive map as we discuss later. Formally, we define the spatial environment as a pair function $(V, \Sigma)$ where $V$ is the set of grid cells and $\Sigma$ is the spatial feature encoded by each grid cell.

\paragraph{Vision:}
Visual cues (\ie landmarks) can help PwD to regain their orientation depending on their intrinsic ability to recognise them \cite{griffith_simulation_2017}. Similarly, upon seeing a disoriented PwD, a nurse can decide whether to intervene or not. To model visual perception for simulated agents, we used a simple ray-casting procedure based on the algorithm described by \citeauthor{amanatides1987fast} (1987). We used parameters \textit{sight} and \textit{field of view (FOV)} where sight is radius of the circle segment that represents agent's attention zone and  FOV is the angle of the circle segment that represents attention zone. FOV is only used for PwD type agent to model an impaired vision.
\paragraph{Agents:}
In the following, we briefly introduce abilities and role of each agent type of agent.
\paragraph{PwD:}
To simulate the behavior of PwD we adapted the cognitive architecture MicroPsi \cite{bach_micropsi_2003} which defines the behavior in terms of three basic drives, namely \textit{Physiological} (\eg sleep, toilet) , \textit{Social} (\ie meeting people to seek affiliation), and \textit{Cognitive} (\eg competence and reduction of uncertainty) desires. We incorporated the desire to reduce uncertainty in our navigation model whereas physiological and social desires are modeled as \textit{needs} along with the desire for competence which is similar to the need of meeting ones appointments (\ie following a schedule). We define \textit{need} as a pair $(\glssymbol{need:requested}, \glssymbol{need:state})$, where \glssymbol{need:requested} \glsdomain{need:requested} \glossentrydesc{need:requested} and \glssymbol{need:state} \glsdomain{need:state} \glossentrydesc{need:state}

The cognitive map, as discussed in the introduction, represents the spatial competence of the PwD agent. We define the cognitive map in terms of the PwD agent's ability to \enquote{memorize} the grid map environment ($V, \Sigma$). Hence the cognitive map is a triple $(\glssymbol{map:workingmemory}, \glssymbol{map:capacity}, \glssymbol{map:pforgetcell})$ where:
\begin{myitemize}
    \item \glssymbol{map:workingmemory} \glsdomain{map:workingmemory} \glossentrydesc{map:workingmemory}
    \item \glssymbol{map:capacity} \glsdomain{map:capacity} \glossentrydesc{map:capacity}
    \item \glssymbol{map:pforgetcell} \glsdomain{map:pforgetcell} \glossentrydesc{map:pforgetcell}
\end{myitemize}

Utilizing the concepts above we define a \textbf{PwD} type agent as an $11$-tuple (\glssymbol{pat:position}, \glssymbol{pat:home}, \glssymbol{pat:schedule}, \glssymbol{pat:pforgetappointment}, \glssymbol{pat:needs}, \glssymbol{pat:cmap}, \glssymbol{pat:state}, \glssymbol{pat:sight}, \glssymbol{pat:fov}, \glssymbol{pat:plandmarks}, \glssymbol{pat:pinterventions}), where:
\begin{myitemize}
    \item \glssymbol{pat:position} \glsdomain{pat:position} and \glssymbol{pat:home} \glsdomain{pat:home} denote agent's current position and set of features that define PwD agent’s home, i.e.,his personal room respectively.
    \item \glssymbol{pat:needs} \glsdomain{pat:needs} \glossentrydesc{pat:needs}
    \item \glssymbol{pat:cmap} \glossentrydesc{pat:cmap} and \glssymbol{pat:state} \glsdomain{pat:state} \glossentrydesc{pat:state}
    \item \glssymbol{pat:sight} \glsdomain{pat:sight} and \glssymbol{pat:fov} \glsdomain{pat:fov} represent PwD agent's \textit{sight} and \textit{field of view} respectively.
    \item \glssymbol{pat:plandmarks} \glsdomain{pat:plandmarks} \glossentrydesc{pat:plandmarks}
    \item \glssymbol{pat:pinterventions} \glsdomain{pat:pinterventions} \glossentrydesc{pat:pinterventions}
\end{myitemize}
To model the PwD agent's navigation we utilized the navigation model proposed in \cite{DP14} which models the navigation process into $4$ steps: $1)$ exploratory process of \textit{perception}, $2)$ strategic process of \textit{goal forming}, $3)$ tactical process of  \textit{route selection}, and $4)$ operational process of  \textit{motion}. In our model, adapting the above-mentioned model, we define the navigation process by two distinct \textit{Act} and \textit{Percept} methods. At each simulation step, a PwD type agent executes the percept/act cycle and tries to navigate towards his current goal. During the \textit{percept} step, an agent perceives his environment and stores features from visible grid cells (\ie within sight \glssymbol{pat:sight} and FOV \glssymbol{pat:fov}) in his cognitive map (\glssymbol{pat:cmap}). If the agent is currently disoriented (\glssymbol{pat:state} = $q_D$), he tries to reset his uncertainty based on visible landmarks (if there are any) with a probability defined by the agent's landmark function (\glssymbol{pat:plandmarks}), and if successful, the agent reorients himself by adding grid cells that represent his current goal into his working memory (\glssymbol{map:workingmemory}). After all visible grid cells have been processed by the agent, the navigation process proceeds with the \textit{act} method to compute the route for agent's current goal. If the agent's current working memory holds all the goal features, state (\glssymbol{pat:state}) is set to be oriented ($q_O$) and a shortest path (from current location to goal) is chosen as a route. However, if the working memory does not contain information about the goal, the state is set to be disoriented ($q_D$) and PwD agent exhibits an \textit{exploration behavior} by visiting the closest grid cells that are not present in his cognitive map at the moment in an attempt to reduce his uncertainty. This navigation behavior is in principle similar to the finding reported by \citeauthor{manning_magellan_2014} (2014).
\begin{figure*}[t]
    \centering
    \begin{subfigure}{0.15\textwidth}
	\centering
	\includegraphics[width=\textwidth]{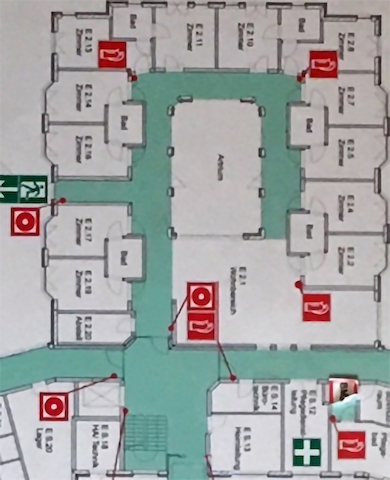}
	\caption{}
	\label{fig:floorplan}
	\end{subfigure}
	\begin{subfigure}{0.15\textwidth}
	\centering
	\includegraphics[width=\textwidth]{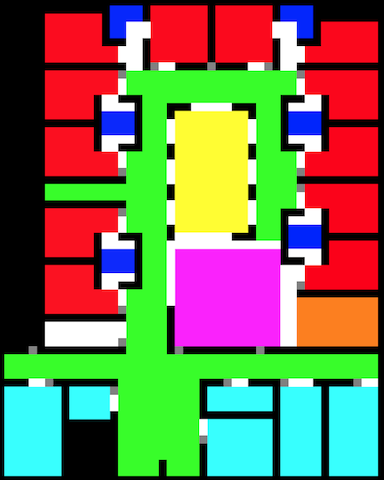}
	\caption{}
	\label{fig:gridmap}
	\end{subfigure}
	\begin{subfigure}{0.16\textwidth}
	\centering
	\includegraphics[width=\textwidth]{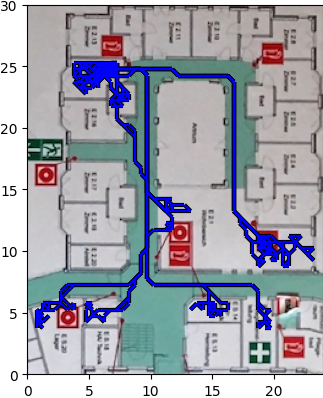}
	\caption{}
	\label{fig:capacity:100}
	\end{subfigure}
	\begin{subfigure}{0.16\textwidth}
	\centering
	\includegraphics[width=\textwidth]{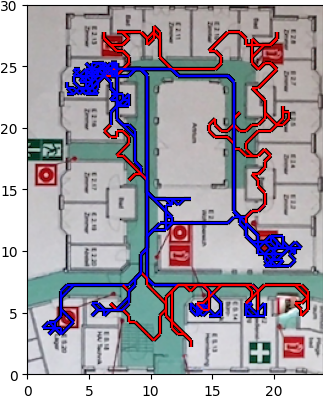}
	\caption{}
	\label{fig:capacity:50}
	\end{subfigure}
	\begin{subfigure}{0.20\textwidth}
	\centering
	\includegraphics[width=\textwidth]{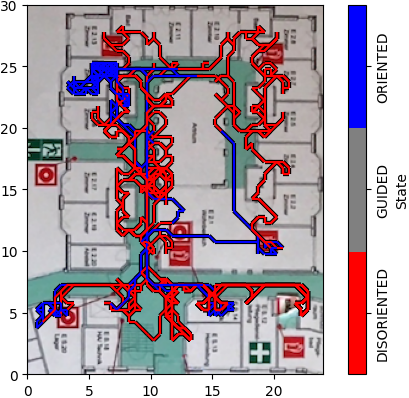}
	\caption{}
	\label{fig:capacity:10}
	\end{subfigure}
    \caption{(a) floor plan of the nursing home, (b) grid image based on (a) : colors represent different aspect of the nursing home such as hallway(green), walls (black), toilet(blue), trajectories of PwD agent having cognitive capacity $100\%$ (c), $50\%$ (d), and $10\%$ (e)}
\end{figure*}
\paragraph{Nurses:}
A nurse agent can guide the PwD agent (when disoriented) towards his current goal. We define a \textbf{nurse} agent as a $5$-tuple (\glssymbol{nur:position}, \glssymbol{nur:home}, \glssymbol{nur:state}, \glssymbol{nur:sight}, \glssymbol{nur:patient}) where:

\begin{myitemize}
    \item \glssymbol{nur:position} and \glssymbol{nur:home} are same as for the PwD type agent.
    \item \glssymbol{nur:state} \glsdomain{nur:state} \glossentrydesc{nur:state}
    \item \glssymbol{nur:sight} \glsdomain{nur:sight} \glossentrydesc{nur:sight}
    \item \glssymbol{nur:patient} \glossentrydesc{nur:patient}
\end{myitemize}
Normally, a nurse agent is in \textit{idle} state ($\glssymbol{nur:state}=q_I$) at her home location (\glssymbol{nur:home}),  until a PwD agent requires assistance. There are two ways to know if assistance is required or not: either a PwD agent must be in nurse's sight (we assume that nurse always knows if a PwD agent is disoriented/oriented) or she get informed by PwD agent's smart watch. Consequently, if assistance is required, nurse starts pursuing the PwD agent ($\glssymbol{nur:state}=q_P$) and takes the lead to guide ($\glssymbol{nur:state}=q_G$) him towards his current goal.

\paragraph{Smart Watch:}
Each PwD type agent can be equipped with a personal smart watch that can assist them in their daily routine. We define the \textbf{smart watch} agent as an $8$-tuple (\glssymbol{watch:patient}, \glssymbol{watch:state}, \glssymbol{watch:cooldown}, \glssymbol{watch:sensormodel}, \glssymbol{watch:nhelp}, \glssymbol{watch:nfail}) where:
\begin{myitemize}
    \item \glssymbol{watch:patient} \glossentrydesc{watch:patient}
    \item \glssymbol{watch:state} \glsdomain{watch:state} \glossentrydesc{watch:state}
    \item \glssymbol{watch:cooldown} \glsdomain{watch:cooldown} \glossentrydesc{watch:cooldown}
    \item \glssymbol{watch:sensormodel} \glsdomain{watch:sensormodel} \glossentrydesc{watch:sensormodel}
    \item \glssymbol{watch:nhelp} \glsdomain{watch:nhelp} \glossentrydesc{watch:nhelp}
\end{myitemize}

At each simulation step, the smart watch monitors the PwD agent and decides whether to trigger one of three (\ie Navigation, Reminder, Help) interventions. A navigation intervention is triggered if the PwD agent is disoriented and if the disorientation is detected by the smart watch based on its sensor model (\glssymbol{watch:sensormodel}). A successful navigation intervention will result in the PwD agent getting reoriented by adding his current goal into his cognitive map. Similarly, a reminder intervention is invoked if the PwD agent forgets a certain appointment. In this case the PwD agent is reminded to go to this particular appointment. After each of these interventions, the smart watch goes into cool-down model (\glssymbol{watch:cooldown}) before the next intervention is triggered. If these interventions show no effect, and number of interventions exceeds the maximum limit (\glssymbol{watch:nhelp}), a help intervention is made and a next available nurse agent closest to the PwD agent is asked to guide the PwD agent and the smart watch goes into waiting mode ($q_W$).
\subsection{Implementation}
\label{sec:implementation}
\model is implemented in Python (3.7) using the agent-based modeling framework Mesa. The spatial structure of the environment is represented as a PNG grid image where each grid cell (\ie pixel) is color-encoded and represents semantic features of the environment (\eg living room, toilet, landmarks etc). Grid image is then represented as a directed graph that includes accessible cells as its vertices and euclidean distance as edge weights. Based on findings by \citeauthor{andresen_wayfinding_2016} (2016), a \enquote{spatial discomfort} penalty is applied by penalizing walking areas that are right next to boundaries and obstacles such that agents keep a realistic distance from them. Path computations are done using the \textit{Floyd-Warshall} algorithm and can be stored for later uses to keep the simulation time minimum. Moreover, agent-specific parameters are stored in JSON files and adding multiple agents in the simulation is quite straightforward. Regarding the evaluation support, simulations based on user-defined parameters and environment can be executed easily. For each simulation; agents' files are generated that include their trajectories, states etc. These files are then used to visualize agents' behavior and interaction in form of plots and videos for later analysis. We implemented \model for an existing nursing home for PwD. Figure 1, shows the floor plan of the nursing home and its transformed grid image.

\subsection{Generalizability}
Even though we designed and developed our system for PwD and nursing homes as primary application domain, it is not limited to it. Every environment which can be abstracted by a grid-based layout of typed cells can be modeled. Those environments can be populated by agents of various types. In this paper, we introduced our cognitive model-based patient agent and simpler agents (nurse and watch). The cognitive model is also not limited to dementia or other forms of cognitive impairments but is inspired by general cognitive architectures based on needs, desires and cognitive maps of the environment to realize plausible behavior simulations.  

\section{Experiments \& Evaluation}
We evaluated the \model model against three cases to simulate navigation behavior with $1)$ no assistance, $2)$ nurse assistance $3)$ nurse and smart-watch assistance. Since there are a large number of possible parameter combinations, we kept the number of variables minimum to report our results in this work. For all experiments, we only change PwD type agent's cognitive capacity within four values (\ie $100\%$, $50\%$, $10\%$), remaining parameters shown in table \ref{tab:params}. Results (table \ref{tab:results}) show that PwD agent's travel efficiency ($TE$), which is the mean ratio of shortest distance to traveled distance, starts decreasing as his cognitive capacity decreases as PwD agent becomes lost more often and his path becomes more tortuous  (note that $TE$ is $0.91$ at $100\%$ capacity instead of $1.0$ because of the discomfort penalty mentioned in section \ref{sec:implementation}). A similar association of level of dementia and its impact on travel efficiency is reported in \cite{martino1991travel}. Moreover,  the number of disorientation episodes ($n$) increase as cognitive capacity decreases. With nurse assistance, $TE$ increases for the same levels of cognitive capacity and the PwD agent starts getting less disoriented or his disorientation episodes become shorter. Similarly, the addition of smart watch agent improves PwD agent's $TE$ and saves nurse agent's time as less human guidance is required. Moreover, the duration of disorientation decreases significantly when using a smart watch as the PwD agent does not only rely on the nurse agent anymore to regain his orientation. 
\begin{table}[b]
\resizebox{\linewidth}{!}{%
\begin{tabular}{@{}llllllllllll@{}}
\toprule
 & \multicolumn{7}{c}{Patient} & \multicolumn{3}{c}{Watch} & Nurse \\ \cmidrule(lr){2-8} \cmidrule(lr){9-11}
 & \glssymbol{pat:schedule} & \glssymbol{pat:pforgetappointment} & \glssymbol{map:pforgetcell} & \glssymbol{pat:sight} & \glssymbol{pat:fov} & \glssymbol{pat:plandmarks} & \glssymbol{pat:pinterventions} & \glssymbol{watch:cooldown} & \glssymbol{watch:sensormodel} & \glssymbol{watch:nhelp} & \glssymbol{pat:sight} \\ \midrule
 & as in sec 3.1 & 0 & 0 & 5 & 90 & 0.1 & 0.8 & 60 & 0.1 & 3 & 10 \\ \bottomrule
\end{tabular}%
}
\caption{Agent-specific parameters for experiments}
\label{tab:params}
\end{table}

\begin{table}[t]
\centering
\resizebox{\linewidth}{!}{%
\begin{tabular}{@{}cclcclllclcclllclcclll@{}}
\toprule
 & \multicolumn{7}{c}{No assistance} & \multicolumn{7}{c}{Nurse assistance} & \multicolumn{7}{c}{Nurse \& Watch assistance} \\ \cmidrule(lr){2-8}\cmidrule(lr){9-15} \cmidrule(l){16-22}

Capacity(\%) & $TE$ & \multicolumn{3}{c}{Disorientation} & \multicolumn{3}{c}{States(\%)} & $TE$ & \multicolumn{3}{c}{Disorientation} & \multicolumn{3}{c}{States(\%)} & $TE$ & \multicolumn{3}{c}{Disorientation} & \multicolumn{3}{c}{States(\%)} \\ \cmidrule(lr){3-5} \cmidrule(lr){6-8} \cmidrule(lr){13-15} \cmidrule(l){17-19} \cmidrule(lr){10-12} \cmidrule(l){20-22}

 &  & $n$ & \multicolumn{2}{c}{episode duration} & $q_O$ & $q_D$ & $q_G$ &  & $n$ & \multicolumn{2}{c}{episode duration} & $q_O$ & $q_D$ & $q_G$ &  & $n$ & \multicolumn{2}{c}{episode duration} & $q_O$ & $q_D$ & $q_G$ \\ \cmidrule(lr){4-5} \cmidrule(lr){11-12} \cmidrule(lr){18-19}

 &  &  & $\mu$ & $\sigma$ &  &  &  &  &  & $\mu$ & $\sigma$ &  &  &  &  &  & $\mu$ & $\sigma$ &  &  &  \\ \midrule
100 & \multicolumn{1}{l}{0.91} & 0 & n/a & n/a & 100 & 0 & 0 & \multicolumn{1}{l}{0.91} & 0 & n/a & n/a & 100 & 0 & 0 & \multicolumn{1}{l}{0.91} & 0 & n/a & n/a & 100 & 0 & 0 \\
50 & \multicolumn{1}{l}{0.85} & 3 & 122 & 21 & 92 & 8 & 0 & \multicolumn{1}{l}{0.86} & 3 & 88 & 51 & 93 & 5 & 2 & \multicolumn{1}{l}{0.90} & 2 & 25 & 28 & 99 & 1 & 0 \\
10 & 0.52 & 25 & 81 & 91 & 58 & 42 & 0 & 0.71 & 22 & 21 & 32 & 77 & 10 & 13 & 0.81 & 29 & 8 & 6 & 85 & 5 & 10 \\ \bottomrule
\end{tabular}%
}
\caption{Simulation results showing assistance effect on PwD agent's spatial orientation. Where $TE$ (Travel Efficiency) is the mean ratio of shortest distance to travelled distance, $n$ is number of disorientation episodes and $\mu$ and $\sigma$ are mean and standard deviation of episodes.}
\label{tab:results}
\end{table}


\section{Conclusion and Future Work}
\model offers a potential way-forward in providing synthetic data on PwD's navigation behavior and their response towards assistance that is essential for modeling and validating assistive technologies. One limitation of our model is the assumption of perfectly compliant PwD who do not oppose their schedule and always comply with the nurse agent, therefore we plan to introduce non-compliant PwD in the future. Moreover, comparison against ground truth data is also essential and planned for the future to validate our model.
\section*{Acknowledgements}
This research is funded by the German Federal Ministry of Education and Research (BMBF) as part of the EIDEC project (01GP1901C), and by the European Union (EU, EFRE) as part of the SAMi project (TBI-1-103-VBW-035).
\bibliographystyle{named}
\bibliography{bib}
\end{document}